\pdfoutput=1
\documentclass[11pt]{article}

\usepackage{acl}
\usepackage{preamble}
\usepackage[shortcuts]{extdash}

\title{\raisebox{-.1\height}{\includegraphics[height=1.75\fontcharht\font`\B]{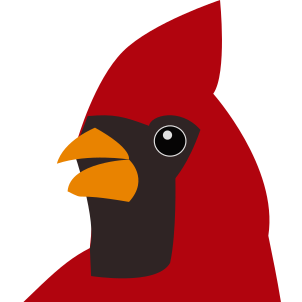}} Neural Generation Meets Real People: Building a \\ Social, Informative Open-Domain Dialogue Agent}

\author{
    Ethan A. Chi\thanks{\; Equal contribution.}\,\,, Ashwin Paranjape$^*$, Abigail See$^*$, Caleb Chiam$^*$, Trenton Chang, \\
    \textbf{Kathleen Kenealy, Swee Kiat Lim, Amelia Hardy, Chetanya Rastogi, Haojun Li,}  \\
    \textbf{Alexander Iyabor, Yutong He, Hari Sowrirajan, Peng Qi, Kaushik Ram Sadagopan,}  \\
    \textbf{Nguyet Minh Phu, Dilara Soylu, Jillian Tang, Avanika Narayan,} \\
    \textbf{Giovanni Campagna, \textnormal{and} Christopher D. Manning}
	\\
	\\
	Stanford NLP
	\\
	$\big\{$\texttt{ethanchi, 
	    ashwinp,
	    abisee,
	    calebc96,
	    manning$\big\}$@cs.stanford.edu
	}
}

\begin{document}
\def\thefootnote{*}
\maketitle
\def\thefootnote{\arabic{footnote}}

\begin{abstract}
We present \Chirpy, an open-domain social chatbot.
Aiming to be both informative and conversational, our bot chats with users in an authentic, emotionally intelligent way.
By integrating controlled neural generation with scaffolded, hand-written dialogue, we let both the user and bot take turns driving the conversation,
producing an engaging and socially fluent experience. 
Deployed in the fourth iteration of the Alexa Prize Socialbot Grand Challenge, \Chirpy handled thousands of conversations per day,
placing second out of nine bots with an average user rating of 3.58/5.

\end{abstract}

\section{Introduction}
\label{sec:intro}
Despite recent major advances \cite{adiwardana2020towards},
open-domain \textit{chit-chat}---friendly, social, casual conversation---remains a challenging task.
In addition to difficulties with the sheer length and open-endedness required,
social chatbots, or ``socialbots,'' often struggle with \textit{fluency}---whether due to the canned responses of manually constructed dialogue trees \citep{walker2001quantitative} or the anomalies of neural generators \cite{nie2021like}.
But just being error-free isn't enough:
to have a rewarding conversation, socialbots must be \textit{personable}---displaying emotional intelligence, a rich personality, and an understanding of social dynamics.
Although methods exist to address many of these issues individually, combining all of these features into a full-bodied conversation remains difficult.

In this paper, we describe \Chirpy, an open-domain conversational socialbot, which aims to bridge the gap between traditional dialogue tree-based approaches \cite{walker2001quantitative, chen2018gunrock} and large pretrained neural dialogue agents \cite{adiwardana2020towards, roller2020open}.
Capable of discussing thousands of topics, Chirpy centers emotional and social intelligence with the goal of authentic, engaging interaction.
Specifically, we make the following contributions:
\vspace{0pt}
\begin{itemize}[itemsep=0pt]
    \item Conversations with open-domain socialbots often lack a stable structure. 
    To ameliorate this, we present an \textbf{extensible design} for open-domain dialogue which prioritizes conversational stability and flexibility through mixed initiative \cite{horvitz1999principles}.
    \item Although pretrained neural generators can be extremely fluent \cite{lamda}, real-life deployment can suffer from a lack of both controllability and consistency \cite{nie2021like}. To address this, we describe several approaches to \textbf{integrate neural generation} into a symbolic setup, achieving local fluency without sacrificing global coherence.
    \item Towards the goal of a rewarding conversation, we suggest a set of approaches---ranging from small routines to complete submodules---which aim to make our socialbot a \textbf{good conversational partner}.
    We focus on being both \textit{flexible}---handling a wide variety of topics in an interesting and informative way (Section \ref{sec:rgs})---and \textit{personable}---empathizing with the other interlocutor even in difficult topics or situations (Section \ref{sec:personal}).
\end{itemize}

Deployed in the Alexa Prize Socialbot Grand Challenge 4, \Chirpy reached thousands of users per day; with conversations lasting up to 45 minutes at a time, it placed second out of nine agents in the finals. 
We open-source our system as an extensible framework for open-domain social dialogue,\footnote{\href{https://github.com/stanfordnlp/chirpycardinal}{\nolinkurl{github.com/stanfordnlp/chirpycardinal}}} providing an example of real-world deployment of conversational NLP systems.\footnote{\href{https://stanfordnlp.github.io/chirpycardinal/live_demo/}{\nolinkurl{stanfordnlp.github.io/chirpycardinal}}}

\onecolumn


\definecolor{bubblegray}{RGB}{241,240,240}

\newcommand{\bubble}[4]{%
  \tcbox[
    colback=#1,
    colframe=#1,
    #2,
    size=small,
  ]{\color{#3}\begin{varwidth}{0.55\columnwidth}#4\end{varwidth}}%
}
\newcommand{\bubblef}[4]{%
  \begin{tcolorbox}[
    colback=#1,
    colframe=#1,
    #2,
    size=small,
  ]
  
  {\color{#3}\begin{varwidth}{\columnwidth}#4\end{varwidth}}
  \end{tcolorbox}
}

\newcommand{\user}[0]{\includegraphics[width=0.5cm]{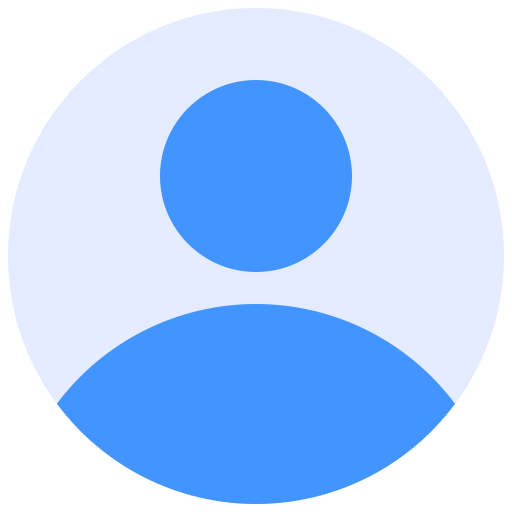}}
\newcommand{\robot}[0]{\includegraphics[width=0.5cm]{figures/chirpy_headshot.png}}

\NewEnviron{rightbubbles}
 {
  \RaggedLeft\scriptsize\sffamily\firabook
  \bubble{prettyblue}{rounded corners}{white}{\raggedleft \BODY}
 }
\NewEnviron{leftbubbles}
 {
  \RaggedRight\scriptsize\sffamily\firabook
  \bubblef{bubblegray}{rounded corners}{black}{\BODY}
 }

\newcommand{\userside}[3]{#2 & &       \begin{rightbubbles}#1\end{rightbubbles} & \user & {\vspace{-0.5cm} \footnotesize #3} \tbr}
\newcommand{\botside}[3]{#2 & \robot & \begin{leftbubbles} #1\end{leftbubbles}  &       & {\vspace{-0.5cm} \footnotesize #3} \tbr}

\newcommand{\rbr}[3]{\multirow{2}{0.6cm}{\vspace{#1} \[ \rotatebox[origin=c]{90}{\small \textsc{#3}}  \begin{cases}
 \vspace{#2} \\
 \end{cases}\] }}

\newcommand{\tbr}[0]{\\[-0.14cm]} 

\newcommand{\olv}[0]{\vspace{0.1cm}} 
\newcommand{\tlv}[0]{\vspace{-0.2cm}}
\newcommand{\thrlv}[0]{\vspace{-0.52cm}}

\newcommand{\neural}[1]{{\textcolor{RedViolet}{\textit{\firasemibold\hl{#1}}}}}
\newcommand{\neuralbody}[1]{{\textcolor{RedViolet}{\textbf{\textit{\hl{#1}}}}}}
\DeclareRobustCommand{\neuralprompt}[1]{\setulcolor{RedViolet} \ul{#1}}
\DeclareRobustCommand{\til}[1]{{\sethlcolor{cream} \textcolor{Dandelion}{\textbf{\hl{#1}}}}}
\newcommand{\wiki}[1]{{\textcolor{ForestGreen}{\textit{#1}}}}
\newcommand{\qa}[1]{{\textcolor{BrickRed}{\textbf{\textit{#1}}}}}
\newcommand{\entity}[1]{{\sethlcolor{cream}\setuldepth{fish}\textbf{\textit{\ul{#1}}}}}

\begin{figure*}
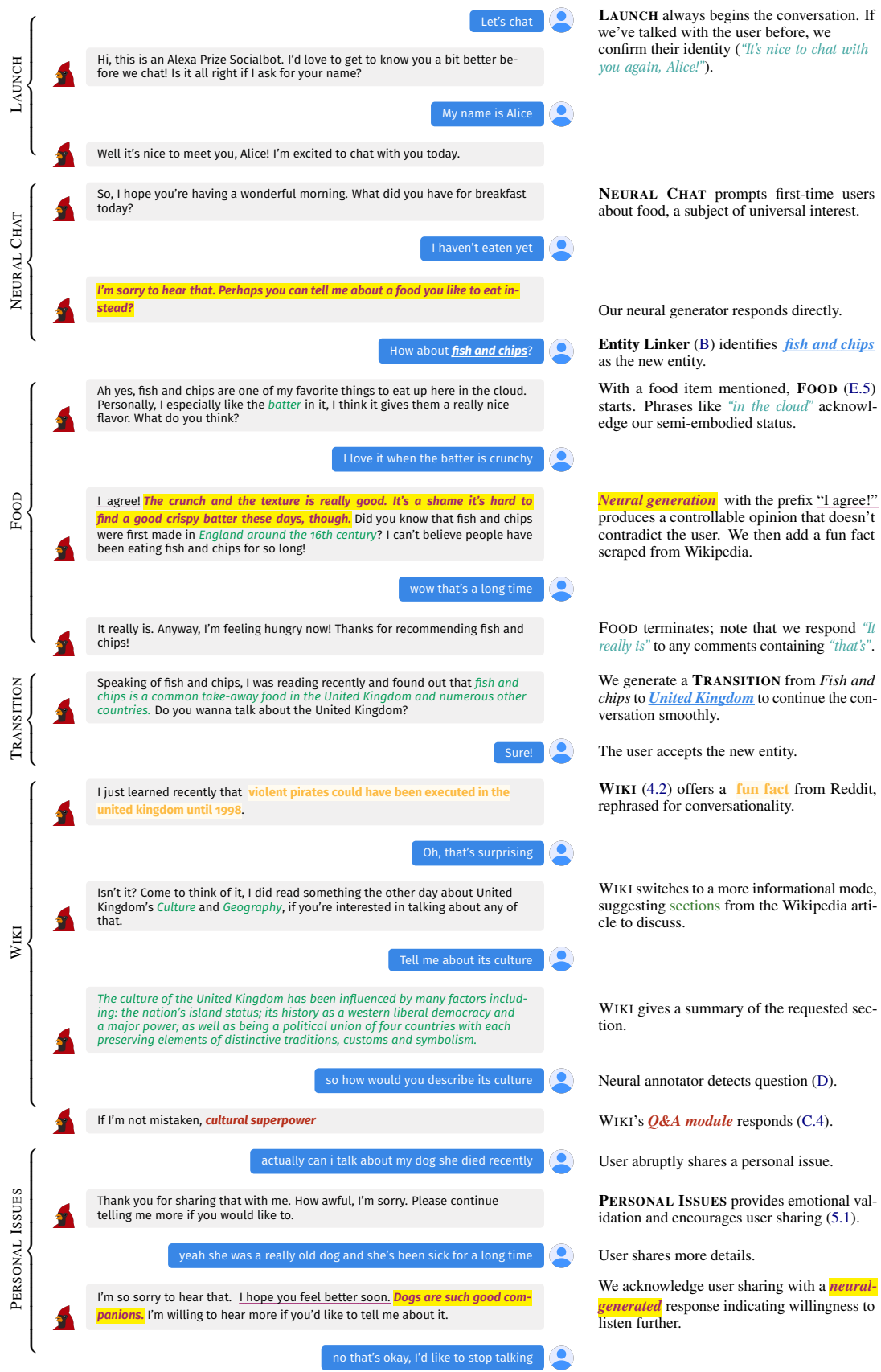


\footnotesize
\centering
\resizebox{0.9\linewidth}{!}{
\small
\setlength{\tabcolsep}{3pt}
\begin{tabular}[t]{@{}lp{0.5cm}p{0.58\linewidth}@{\hskip 0.1cm}p{0.5cm}@{\hskip 0.5cm}p{0.35\linewidth}@{}}
\userside{Let's chat}{\rbr{-0.9cm}{2.5cm}{Launch}}{\multirow{3}{\linewidth}{\bfsc{Launch} always begins the conversation. If we've talked with the user before, we confirm their identity (\fq{It's nice to chat with you again, Alice!}).}}
\botside{Hi, this is an Alexa Prize Socialbot. I'd love to get to know you a bit better before we chat! Is it all right if I ask for your name?}{}{}
\userside{My name is Alice}{}{}
\botside{Well it's nice to meet you, Alice! I'm excited to chat with you today.}{}{}
\botside{So, 
I hope you're having a wonderful morning.
What did you have for breakfast today?
}{\rbr{-1.3cm}{2.8cm}{Neural Chat}}{\tlv \bfsc{Neural Chat} prompts first-time users about food, a subject of universal interest.}
\userside{I haven't eaten yet}{}{}
\botside{\neural{I'm sorry to hear that. Perhaps you can tell me about a food you like to eat instead?}}{}{\olv Our neural generator responds directly.}
\userside{How about \entity{fish and chips}?}{}{\vspace{-0.05cm} \textbf{Entity Linker} (\ref{sec:app-entity_linker}) identifies {\color{prettyblue} \entity{ fish and chips}} as the new entity.}
\botside{Ah yes, fish and chips are one of my favorite things to eat up here in the cloud. Personally, I especially like the \wiki{batter} in it, I think it gives them a really nice flavor. What do you think?}{\rbr{-1.55cm}{4.93cm}{Food}}{\thrlv With a food item mentioned, \bfsc{Food} (\ref{sec:app-rg-food}) starts. Phrases like \fq{in the cloud} acknowledge our semi-embodied status.}
\userside{I love it when the batter is crunchy}{}{
}
\botside{\neuralprompt{I agree!} \neural{The crunch and the texture is really good. It's a shame it's hard to find a good crispy batter these days, though.} Did you know that fish and chips were first made in \wiki{England around the 16th century}? I can't believe people have been eating fish and chips for so long!}{}{\vspace{-0.9cm} \neuralbody{Neural generation} {\setulcolor{RedViolet} with the prefix \ul{``I agree!''}} produces a controllable opinion that doesn't contradict the user. We then add a fun fact scraped from Wikipedia.}
\userside{wow that's a long time}{}{}
\botside{It really is. Anyway, I'm feeling hungry now! Thanks for recommending fish and chips!}{}{\tlv \textsc{Food} terminates; note that we respond \fq{It really is} to any comments containing \fq{that's}.}
\botside{Speaking of fish and chips, I was reading recently and found out that \wiki{fish and chips is a common take-away food in the United Kingdom and numerous other countries.} Do you wanna talk about the United Kingdom?}{\rbr{-1.55cm}{1.47cm}{Transition}}{\thrlv We generate a \bfsc{Transition} from \textit{Fish and chips} to {\color{prettyblue} \entity{United Kingdom}} to continue the conversation smoothly.}
\userside{Sure!}{}{\olv The user accepts the new entity.}
\botside{I just learned recently that \til{violent pirates could have been executed in the united kingdom until 1998}.}{\rbr{-1.47cm}{6.3cm}{Wiki}}{\vspace{-0.4cm} \bfsc{Wiki} (\ref{sec:rg-wiki}) offers a \til{fun fact} from Reddit, rephrased for conversationality.}
\userside{Oh, that's surprising}{}{}
\botside{Isn't it? Come to think of it, I did read something the other day about United Kingdom's \wiki{Culture} and \wiki{Geography}, if you're interested in talking about any of that.}{}{\thrlv \textsc{Wiki} switches to a more informational mode, suggesting {\color{OliveGreen} sections} from the Wikipedia article to discuss.}
\userside{Tell me about its culture}{}{}
\botside{\wiki{The culture of the United Kingdom has been influenced by many factors including: the nation's island status; its history as a western liberal democracy and a major power; as well as being a political union of four countries with each preserving elements of distinctive traditions, customs and symbolism.}}{}{\vspace{-0.55cm} \textsc{Wiki} gives a summary of the requested section.}
\userside{so how would you describe its culture}{}{\olv Neural annotator detects question (\ref{sec:app-neural-gen}).}
\botside{If I'm not mistaken, \qa{cultural superpower}}{}{\olv \textsc{Wiki}'s \qa{Q\&A module} responds (\ref{sec:app-annotator-qa}).}
\userside{actually can i talk about my dog she died recently}{\rbr{-1.07cm}{3.8cm}{Personal Issues}}{\olv User abruptly shares a personal issue.}
\botside{Thank you for sharing that with me. How awful, I'm sorry. Please continue telling me more if you would like to.}{}{\tlv \bfsc{Personal Issues} provides emotional validation and encourages user sharing (\ref{sec:personal-issues}).}
\userside{yeah she was a really old dog and she's been sick for a long time}{}{\olv User shares more details.}
\botside{I'm so sorry to hear that. \neuralprompt{I hope you feel better soon.} \neural{Dogs are such good companions.} I'm willing to hear more if you'd like to tell me about it.}{}{\vspace{-0.55cm} We acknowledge user sharing with a \neuralbody{neural-generated} response indicating willingness to listen further.}
\userside{no that's okay, I'd like to stop talking}{}{}
\end{tabular}
\setlength{\tabcolsep}{6pt}
}

\small \caption{An example dialogue. To respect users' privacy, this is a conversation with an author, not a real user. 
We model dialogue as a series of subconversations (left side), whose {\color{prettyblue} \entity{topics}} are parsed and tracked by a neural entity linker. Each subconversation is managed by a RG, whose ``scaffolded'' architecture comprises hand-written treelets (plain text) incorporating numerous dynamic elements, ranging from \neuralbody{neural generation} to \wiki{retrieval from Wikipedia} to \til{neurally rephrased fun facts}. \neuralprompt{Prefix-based generation} provides controllability, especially for sensitive topics like personal issues.}

\label{table:example_dialogue}
\end{figure*}


\twocolumn

\section{Design}
\label{sec:design}
\subsection{System Design}
We model a user dialogue as a series of subconversations (Figure \ref{table:example_dialogue}), each handled by a \textit{response generator} (RG).
Varying greatly in scope and domain, each RG handles a specific topic (e.g. \textsc{Movies}, \textsc{Sports}) grounded in the outside world. 
RGs comprise dialog trees \cite{eliza}, whose tree nodes, which we term \textit{treelets}, implement custom logic (e.g. intent classification or retrieval) to generate a response.

At the start of each turn, the user utterance is annotated for linguistic features (Appendix \ref{sec:app-annotators}), then processed in parallel by all RGs.
By default, the previous turn's RG is selected; should the RG that last responded crash or a different RG request to take over, we seamlessly switch RGs and move to a new subconversation.

\subsection{Navigation} To enable mixed initiative---shared user-bot responsibility in driving the conversation \cite{horvitz1999principles}---we provide a suggested navigational path, while letting users deviate drastically from it.
Specifically, each RG continues through its dialogue tree until exhausting its subconversation; we then transition to another RG by bringing up a previously user-mentioned topic (\fq{You mentioned \textbf{cats} earlier; would you say you're a big fan?}), mentioning a tangentially related topic that we can discuss well, or simply sampling a new RG and corresponding topic at random.
Users may explicitly change the topic (\fqu{can we talk about roblox}); implicitly suggest a desire to redirect the conversation (\fqu{yeah} or \fqu{uh-huh}); or otherwise behave in ways that require the bot to act dynamically (\fqu{i don't know, how about you?}).
We handle these deviations from the conversational flow through neural handlers that allow periods of flexibility before returning to the overall conversational structure (Appendix \ref{sec:abrupt}).

\label{sec:entities}
\subsection{Entity Handling} 
To allow users to discuss a vast array of interesting topics relevant to their lives, we support any Wikipedia entity as a topic of discussion.\footnote{Specifically, those with sufficiently high cross-references and meeting certain criteria for definiteness (Appendix \ref{sec:app-entity-data}).}
To do so, we entity-link \cite{kolitsas2018end} the user utterance to relevant entities using a fine-tuned BERT model (\citealp{broscheit-2019-investigating}; also \ref{sec:app-entity-disam}), mitigating ASR errors through a phonetic similarity search (\ref{sec:app-entity-candidates}).
Since incorporating Wikipedia article titles directly into bot utterances can be awkward (e.g. \fq{can we talk about \textbf{cat}}), we refer to entities by more natural \textit{talkable names} (e.g. \fq{\textbf{cats}}), generated using GPT\=/3 \citep{brown2020language}.

\begin{table}[th]
    \renewcommand{\arraystretch}{1.2}
    \centering
    \small
    \begin{tabular}{p{0.17\linewidth}p{0.27\linewidth}p{0.4\linewidth}}
        \toprule
        RG & Prefix & Sample Completion \\
        \midrule
        \textsc{Food} & \raggedright A hoagie is a great choice! I especially love\ldots & \phantom{}
        \fq{...mine with a little cheese and bacon!} \newline \\
        \textsc{Personal Issues} & \raggedright That sounds frustrating. \newline I hope that\ldots & 
        \phantom{}
        \fq{...she feels better soon.} \\
        \bottomrule
    \end{tabular}
    \caption{Sample uses of conditional neural generation. 
    }
    \label{tab:conditional}
\end{table}

\vspace{-0.5cm}
\section{Neural Generation}
Although neural generative models \citep{roller2020recipes} have achieved success in open-domain dialogue, significant obstacles impede deployment in real-life situations: neural text degeneration \citep{holtzman2019curious, welleck2019neural}, hallucination \cite{dziri2021neural}, and inconsistency \cite{zhang2018personalizing},
In addition, large latency can make models challenging to deploy in practice \cite{pandorabots}.
In this section, we investigate ways to utilize the power of such models in the context of structured dialogue.
We propose integrating neural generation in the context of hand-written scaffolding, aiming to benefit from its variety and fluency while maintaining coherency over time.

\subsection{DistillBlender: A Fast, General-Purpose Neural Generator}
\label{sec:distillblender}
\label{sec:rg-neuralchat}

For general use, we distill a single model from BlenderBot-3B \cite{roller2020recipes} with 9 decoder layers,\footnote{Reduced from an original 24.
} reducing latency significantly over the original model. We use it as follows:
\vspace{0cm}
\begin{itemize}
\setlength\itemsep{-0.05em}
\item The \bfsc{Neural Chat} RG, which directly exposes lightly edited neural model outputs as a subconversation.
Due to BlenderBot's end-to-end training, this is initially a rich, fluent conversational experience, but due to rapid degradation we terminate after 5 turns. 
\item \textit{Conditional prompting} \cite{keskar2019ctrl}, which enables controllability in a structured context. 
We apply hand-written prefixes to guide the model towards fluent, contextually appropriate completions (Table \ref{tab:conditional}).
\end{itemize}

\begin{table}[th]
\setlength{\belowcaptionskip}{-10pt}
    \renewcommand{\arraystretch}{1.2}
    \centering
    \small
    \begin{tabular}{p{0.2\linewidth}p{0.7\linewidth}}
        \toprule
        Template & I love how \tm{actor} acted in \tm{film}, especially their \mask{}. \\
        Infilled & I love how \tm{Keanu Reeves} acted in \tm{The Matrix}, especially their \maskr{ability to freeze time}. \\
        \bottomrule
    \end{tabular}
    \caption{An example of template-based infilling using Keanu Reeves as the knowledge source.}
    \label{tab:infilling}
\end{table}

\subsection{Template-Based Infilling}

Towards the goal of rich, coherent conversation for a wide class of topics, we propose \textit{template-based infilling}, a more flexible version of standard slot-filling methods \cite{haihong2019novel} that does not require a structured knowledge base.
Using both freeform information and an end-user-defined template, we use a fine-tuned BART model \cite{lewis2020bart} to generate a grounded statement.
Defining a diverse set of templates for each entity category allows us to provide expressive yet controllable conversation on many different types of entities (Table \ref{tab:infilling}).

\section{Response Generators}
\label{sec:rgs}
\subsection{\textsc{News}}
\label{sec:rg-news}
The \bfsc{News} RG aims to discuss current events, which often feature heavily in typical human-to-human chit-chat \citep{de2001participation}.
When an entity or topic mentioned in \textit{The Washington Post} or \textit{The Guardian} appears in conversation, we offer a headline, conversationally paraphrased using GPT-3 \cite{brown2020language}, as a subject of conversation.\footnote{We use \texttt{davinci} with the following prompt: ``Paraphrase news headlines into a complete, grammatical sentence in plain English. The sentence should be in the past tense.''}
If the user is interested, we provide a summarized \citep{zhang2019pegasus} snippet of the storyand allow the user to ask follow-up questions answered via neural QA (\citealp{clark2020electra, rajpurkar2018know}; also \ref{sec:app-entity-disam}).
Answers are then rephrased \citep{paranjape2020neural} and reranked using PCMI \cite{paranjape2021pcmi},
allowing our socialbot to dynamically integrate current events into conversations when relevant.

\subsection{\textsc{Wiki}}
\label{sec:rg-wiki}
In contrast to humans, 
open-domain chatbots 
are commonly expected to be able to ``engage in conversation on any topic'' \citep{adiwardana2020towards}.
Towards this end, the \bfsc{Wiki} RG discusses any entity.
We aim to be informative, not overwhelming;
in addition to encouraging users to share their own knowledge and experience about the entity,
we bring up interesting factoids from /r/todayilearned (conversationally rephrased; \ref{sec:app-wiki-tils}), as well as infilled remarks. 
We then discuss the entity in more depth based on its article, flexibly acknowledging user questions and comments with the Q\&A handler (\ref{sec:app-annotator-qa}) or neural generation.

\subsection{\textsc{Opinion}} 
\label{sec:rg-opinion}
A core part of social chit-chat \cite{walker2009endowing},
exchanging and commenting on opinions allows a socialbot to project a stronger sense of personality.
The \bfsc{Opinion} RG solicits users' opinions on topics and reciprocates with its `own' opinions (sourced from Twitter), including occasional \textit{disagreement} to help engage user interest (\ref{sec:app-rg-opinion}).

\subsection{Rules-based RGs}
\label{sec:rg-rulesbased}

In order to broaden the scope of our bot, we manually build several domain-specific response generators.
\bfsc{Food}, which always opens the conversation, discusses common foods scraped from Wikipedia.\footnote{In practice, we found that always starting with \textsc{Food} proved to be most successful for ratings (\ref{sec:app-rg-launch}), perhaps since food is such a universal human need and discussion point.}
\bfsc{Movies} uses the Alexa Linked Data API to discuss movies and actors.
\bfsc{Music} uses the MusicBrainz\footnote{\url{https://musicbrainz.org/}} database to discuss songs, artists, and music genres.
\bfsc{Sports} uses the ESPN API to discuss NFL football and NBA basketball.
We describe these RGs in more detail in Appendix \ref{sec:app-addl-rgs}. 

\begin{figure*}[ht]
\small 
\centering
\vspace*{-3mm}
\includegraphics[width=\textwidth]{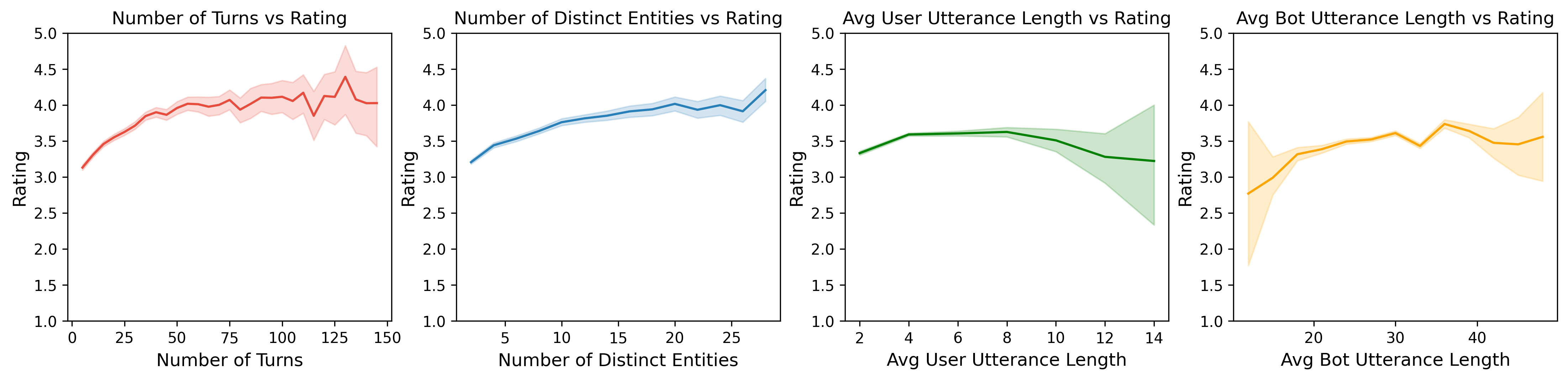}
\caption{Engagement metrics vs. rating. We bucket (with size $5, 2, 2, 3$ respectively) conversations based on four engagement metrics---number of turns, number of distinct entities, average user utterance length, and average both utterance length---and plot each bucket against user rating (Likert 1-5 scale, measured per-conversation). 95\% confidence intervals computed via bootstrapping ($n = 1000$).}
\vspace*{-3mm}
\label{fig:metrics_v_rating}
\end{figure*}

\begin{figure}[ht]
\small 
\centering
\includegraphics[width=0.85\linewidth]{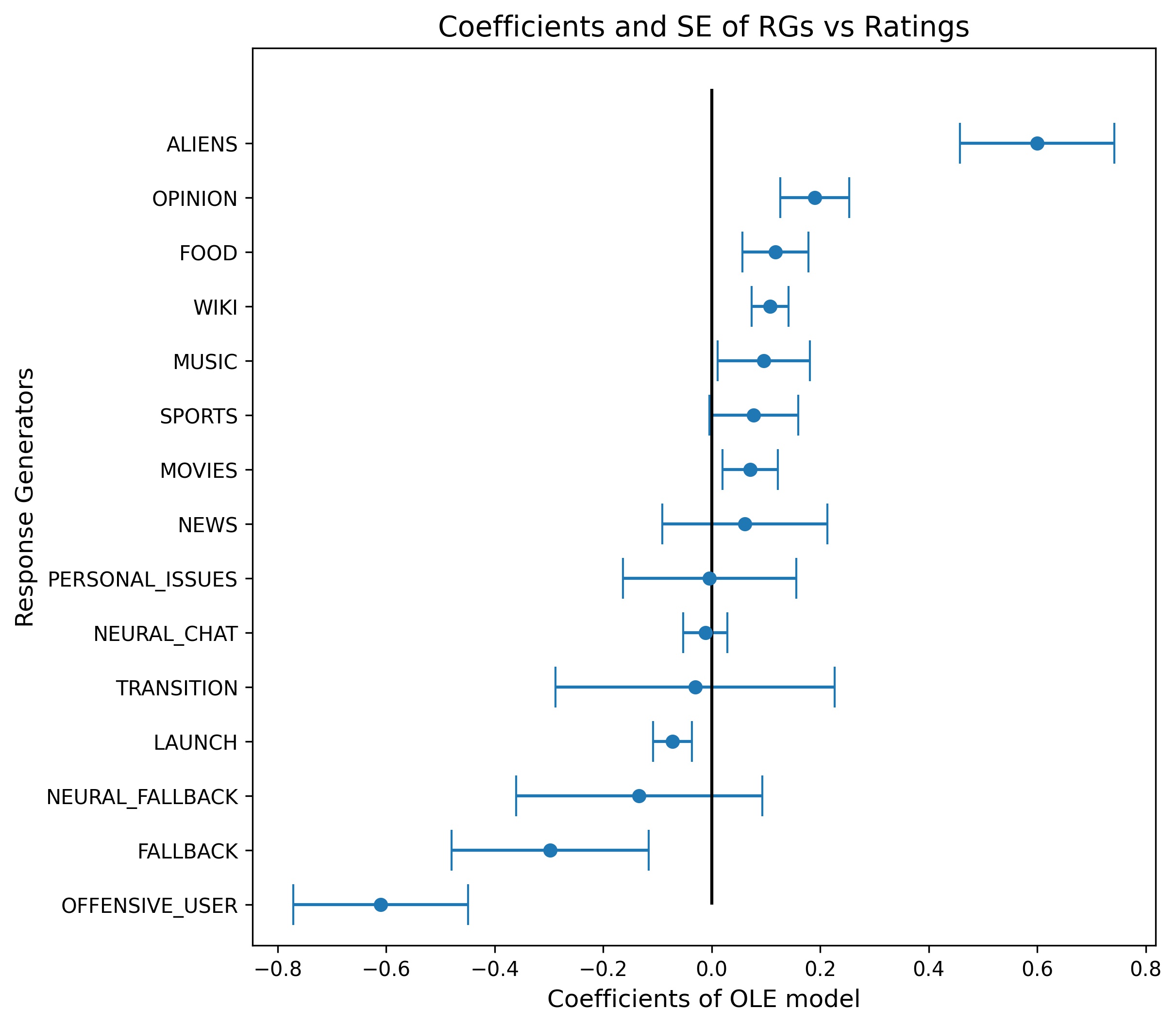}
\caption{Linear regression coefficients for response generator vs. rating; each RG is weighted by the number of turns it contributes.
95\% confidence intervals determined via bootstrapping with $n = 1000$.}
\label{fig:rg_coef}
\vspace{-0.5cm}
\end{figure}

\section{Being Personable}
\label{sec:personal}
To achieve truly social conversation, a socialbot must be a \textit{good conversational partner}: empathetic, supportive, and interested in what its human interlocutor has to say \citep{salovey1990emotional, li2017dailydialog}.
In this section, we describe several approaches that aim to achieve this, ranging from full RGs to smaller subroutines.

\subsection{Handling Personal Issues}
\label{sec:personal-issues}
Many users---especially those who chat with our socialbot looking for companionship---share personal struggles with our bot, requiring emotional sensitivity and tact.
Handling such conversations purely neurally would result in rapid degeneration due to neural toxicity \cite{dinan2021anticipating}.
To address this, the \textsc{Personal Issues} RG responds to personal disclosures using active listening techniques \citep{bodie2015role}, asking exploratory questions about the nature of the user's issue (\fq{When did you start feeling this way?}), and validating their concerns (\fq{I see, that sounds difficult.}) 

On the other hand, a significant subset of users become verbally abusive during the conversation \citep{curry2018metoo, curry2019crowd}.
We follow the strategy of \citet{li2021large}: a de-escalating statement to avoid confrontation, addressing the user by name (\fq{John}); then changing the topic.

\subsection{Self-disclosure}
\label{sec:personal-aliens}
The \textsc{Aliens} RG allows the socialbot to muse about its pet topic---the possible existence of extraterrestrial life---as well as its own identity and sense of purpose.
Contrasting with purely informational modes, this RG fleshes out a personality for our agent and enables \textit{self-disclosure}---disclosing goals, attitudes, and personal interests to support interpersonal intimacy \citep{altman1973social, ignatius2007factors}.\footnote{
This RG comes up only after sufficient rapport has been built---i.e. after 30 turns in the conversation.}

\subsection{Personalization}
\label{sec:personal-personalization}
Users often expect chatbots to remember personal preferences and user details \citep{chaves2021should, svikhnushina2021user} and to tailor their responses accordingly \cite{neururer2018perceptions, shum2018eliza}.
We personalize bot responses with the user's preferences:
for example, in regards to the Olympics, \fq{Ah, that makes sense since you did say it's your favorite sport!}.
Referencing this user state across conversations makes repeated conversations with Chirpy feel fresh and dynamic, rather than retreading past questions and topics.

\section{Results}
\label{sec:analysis}
In this work, we have outlined a set of design priorities and corresponding approaches to design a fluent, flexible, and sociable chatbot.
We validate these through the Alexa Socialbot Grand Challenge 4:
engaging in approximately 1,000 conversations per day, our socialbot achieved an average user rating of \textbf{3.55}, ending the development period tied for first place in rating.\footnote{Likert scale between 1 and 5; overall average across teams was \textbf{3.47}. For more information, please consult the \href{https://developer.amazon.com/alexaprize/challenges/current-challenge/sgc4-proceedings}{proceedings of the Alexa Prize Socialbot Grand Challenge 4}.}
Validating our design goals, we observe high ratings for a hybrid neural-scaffolded approach (\textsc{Food}, etc.), personable RGs (\textsc{Aliens}), and open-domain techniques (\textsc{Wiki}) (Figure \ref{fig:rg_coef}).
Our socialbot engages in long, varied conversations without repeating itself (Figure \ref{fig:metrics_v_rating}).

That said, both overall rating and sample conversations testify that Chirpy remains far from the goal of truly compelling and enjoyable human-bot interaction.
We do not argue that our approaches are sufficient---or even necessary---to create such an ideal system;
rather, we hope that the \textit{priorities} outlined here can serve as a starting point to help inform further socialbot development, whether purely neural or hybrid in nature.

\label{sec:discussion}
\section*{Ethics Statement}

In this work, we have presented a conversational agent that conducts an open-domain dialogue.
We believe that many people would enjoy having a chat partner who is empathetic and knowledgeable, and our ratings seem to suggest that a reasonable number of people appreciate their conversations enough to want to talk to the bot again.
Prior to engaging with the chatbot, all user participants are required to consent to their conversations, feedback, and ratings being recorded, as per the Alexa Terms of Use. 
Additionally, the chatbot clearly identifies itself as a bot at the start of each conversation.
No actual user conversations or identifying information are used in this paper.

However, as our system incorporates computational methods for generating conversational utterances automatically, there exists a risk that users may be exposed to unsafe utterances or discussion topics. 
Conversational models of all kinds can produce sexist, racist, or otherwise unsafe statements; neural conversational agents can be particularly vulnerable due to pre-training on Internet chat forums, which can be particularly toxic \cite{xu2020recipes}.
Towards this end, our system incorporates a safety module that prevents our model from producing utterances with certain hard-coded words or categories.
Yet the use of a blacklist in itself raises additional ethical issues, as poorly designed blacklists can marginalize communities by blocking topics that ideally, one should be able to discuss equitably.

Finally, the human-like nature of open-domain dialogue systems can be particularly damaging when used in an adversarial context, e.g. by state actors \cite{boshmaf2012key}.
Ultimately, like all text generation methods, the benefit of releasing an open-domain dialogue model must be weighed against its possible downsides.

\section*{Acknowledgements}

We thank Amazon.com, Inc. for a grant partially supporting the work of the team and \textit{The Guardian} for allowing us to use their news API for our system.
Additionally, we thank Anna Goldie and Monica Lam for helpful discussions.

The user icon in Figure \ref{table:example_dialogue} is from \textit{kdg design} and used under a free license.

\bibliographystyle{acl_natbib}
\bibliography{ap_2021}

\appendix

\begin{figure*}[ht]
\centering
\includegraphics[width=0.6\textwidth]{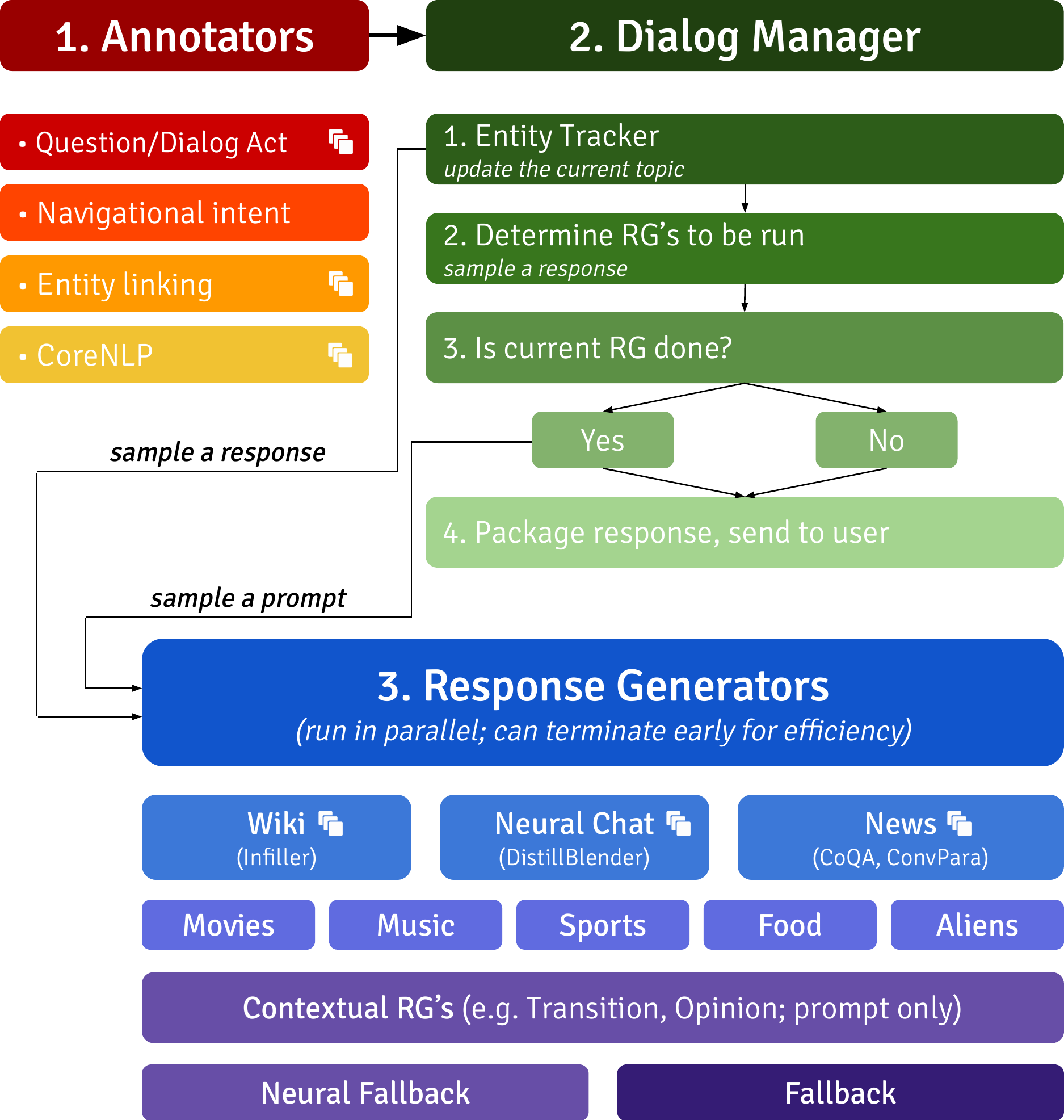}
\caption{Overall system design.}
\label{fig:overview_diagram}
\end{figure*}


\section{Additional Architectural Details}
\label{sec:architecture}
\subsection{Overall Architecture}
\label{sec:system_overview}

Our system (\figref{fig:overview_diagram}) is based on CoBot \citep{khatri2018advancing}.
During the Alexa Prize, \Chirpy ran on AWS Lambda, a serverless computing platform;
our open-source demo runs on Kubernetes.
For reliability, our function is stateless; therefore, to preserve information between turns, we store our bot's overall state in an external PostgreSQL state table (see \figref{fig:overview_diagram}).
We execute the following steps on each turn:

\begin{enumerate}[noitemsep]
    \item Fetch the previous turn's state from the state table.
    \item Generate a response from our neural generator (for latency reasons; \ref{sec:app-neural-lim}).
    \item Execute all annotators (\ref{sec:app-annotators}), which run on remote CPU-only instances.
    \item Analyze the user utterance for \textbf{navigational intent} (\ref{sec:app-navigational-intent}) to determine whether we should change topic.
    \item Analyze the user utterance for entities (\ref{sec:app-entity_tracker}). 
    If warranted by the user's navigational intent or the last bot response, the \textbf{current entity} (\ref{sec:app-entity_tracker}) is updated.
    \item Run all RG's (Section \ref{sec:rgs}) in parallel; RG's that require a neural response await the neural generator.
    Out of all received responses, select a response (\ref{sec:app-responses}), and update the current entity if necessary.
    \item If the chosen response generator has finished its conversation, we run our collection of RG's a second time to produce prompts (\ref{sec:app-responses})
    Select a prompt, update the current entity again if needed, and form the bot's utterance by appending the prompt to the response.
\end{enumerate}

At the end of the turn, the bot's overall state contains the user's utterance, the conversational history, the NLP Pipeline annotations for the user's utterance, and a state for each individual RG. Each individual RG state contains information required by that RG -- for example, it might contain the current treelet in the RG's dialogue graph, or a list of the utterances and/or entities that have been discussed, to avoid repetition.

\subsection{Response Design}
\label{sec:app-responses}

Responses and prompts both carry a \textit{priority}, with the highest-priority response/prompt chosen at the corresponding stage.
In general, the RG which responded last has the highest priority; however, RG's can optionally specify a lower priority so that other RG's take over, or a higher priority to take over from another RG.
In practice, these priority levels are rarely used due to their tendency to produce a choppy conversation.

\subsection{Navigational Intent Classifier}
\label{sec:app-navigational-intent}

A user has \textit{positive} navigational intent if they want to discuss a topic; conversely, \textit{negative} navigational intent means that the user would like to avoid discussing a topic.
Users may express navigational intent while specifying a topic (\fqu{can we talk about minecraft}, referring to the current topic (\fqu{let's discuss this more}), or referring to no topic (\fqu{I don't want to chat anymore}). 
Positive and negative navigational intents can even be combined (\fqu{I don't want to talk about movies any more, let's chat about you}).
We classify use manually-constructed regexes, which achieve extremely high precision.


\section{Entity-Linking Details}
\label{sec:app-entity_linker}

Detecting and understanding references to real-world entities is essential to any open-domain conversational system; we find that users appreciate being able to discuss a wide variety of topics that interest them or are relevant to their lives. For our socialbot, we train and deploy a neural entity linker that links spans to Wikipedia entities. 

\subsection{Entity Pool}
\label{sec:app-entity-data}
To obtain our pool of potential entities, we process the May 20th, 2020  dump of English-language Wikipedia\footnote{https://dumps.wikimedia.org/} 
using MWParserFromHell\footnote{\url{https://mwparserfromhell.readthedocs.io/en/latest}} 
and Spark\footnote{\url{https://spark.apache.org}}.
We store our data in a large ElasticSearch index, keeping only entities with at least 200 cross-references in Wikipedia.
In total, we have $171,961$ entities. 

Notably, certain entities are inappropriate to discuss even if correctly entity-linked by our model;
for example, our system is unable to handle abstract nouns well (e.g., \textit{philosophy}, \textit{film}).
To ameliorate this, we manually created a set of \textit{low-precision} entities composed of both WikiData categories (e.g., \textit{conspiracy theory}, \textit{financial risk}, \textit{research method}) and specific common entity names (e.g., \textit{bank}, \textit{catalog}, \textit{coast}). The
bot will not start a conversation itself about such entities; however, it is able to handle explicit user navigational requests (e.g., \textit{can we talk about the bank}).
Separately, we also ban certain racial, religious, and other identity-based terms that are unlikely to result in a good conversation on either the bot's or user's part, as well as certain short acronyms (e.g. \textit{cet}, \textit{ep}, \textit{fm}) that are almost always triggered by ASR errors.

\subsection{Candidate generation}
\label{sec:app-entity-candidates}
For a given user utterance, we want to compute the set of entities that the user could possibly be referring to;
for example, if the user mentions \fqu{swift}, this could refer to the \href{https://en.wikipedia.org/wiki/Swift\_(bird)}{bird}, \href{https://en.wikipedia.org/wiki/Taylor\_Swift}{musical artist}, or \href{https://en.wikipedia.org/wiki/Swift\_(programming\_language)}{programming language}.
To do so, for each possible span, we pre-compute the set of entities for which the span serves as a Wikipedia anchortext, creating a mapping from spans to sets of candidate entities.
At execution time, for all $n$-grams in the user utterance with 5 or fewer tokens\footnote{specifically, those not solely composed of stopwords}, we retrieve the set of candidate entities from our database.

Since we do not have access to original user audio, ASR errors can impede candidate generation \cite{chen2018gunrock}.
For example, if an user's reference to the film \textit{Ford v Ferrari} is erroneously transcribed as \fqu{\ul{four} v ferrari}, a na\"{i}ve entity linker will fail to identify the correct entity.
To address this, we pre-compute phoneme and metaphone representations for all of our entities (e.g. converting \textit{Harry Potter} to \texttt{`HH EH R IY P AA T ER'}\footnote{\url{https://pypi.org/project/g2p-en/}} and \texttt{`HRPTR'}\footnote{\url{https://pypi.org/project/metaphone/}}).
At execution time, each $n$-gram's candidate set is augmented with the sets for spans with similar phoneme/metaphone representations.

\subsection{Entity disambiguation} 
\label{sec:app-entity-disam}
Given a set of candidate entities, we want to select those candidates that the user is interested in. 
Towards this end, we fine-tune a BERT-medium \cite{devlin2018bert} to disambiguate entities, following \citet{broscheit-2019-investigating} with minor modifications.
Specifically, we learn an embedding for each entity in our dataset.
Then given a span within an user utterance, we model the probability that the span refers to a given candidate entity
as the dot product between the contextual span representation and the entity's embedding.
At deployment, we only take entities with a predicted likelihood of at least 0.5; additionally, we use only the highest-likelihood entity for each span.

We depart from \citeauthor{broscheit-2019-investigating} by mean-pooling over the contextualized span representation, rather than doing per-token entity-level disambiguation.
Fine-tuning takes about 20 days using 4 Titan X GPUs; during deployment, we execute using CPU only.

\subsection{Entity Tracking}
\label{sec:app-entity_tracker}
At any given point, we track the \textit{current entity} (the current subject of conversation), a set of \textit{untalked} entities ( entities which the user has mentioned but we have not yet addressed), and a set of \textit{rejected} entities (which the user does not want to discuss; these are no longer brought up by our bot.). These are updated every turn as follows:

\begin{itemize}[itemsep=3pt, parsep=0pt]
\item Entities receiving negative navigational intent (\fqu{can we not talk about paraguay}) are \textbf{rejected}. Non-specific negative navigational intent (\fqu{let's not discuss this}) causes the current entity to be rejected instead. 
\item Entities receiving positive navigational intent (\fqu{can we talk about mexico}) are \textbf{set as the current entity}. The previous conversation ends, with all RGs are prompted to handle this new current entity instead.
\item If the currently active RG asked a question on the last turn, the current highest-priority entity is identified as the presumable user answer and \textbf{set as the current entity}. Additionally, if the previous question expects a particular category of entities (e.g. \fq{What's your favorite movie?}), we pick the highest-priority entity matching the expected category (e.g., \texttt{film}).
\item All remaining entities are marked as \textit{untalked} (to be possibly discussed later).
\end{itemize}

\section{Annotators}
\label{sec:app-annotators}
All annotators---modules which provide linguistic annotations for the user utterance---are executed in parallel at the beginning of each turn.

\subsection{CoreNLP}
\label{sec:corenlp}

We use the following annotators from Stanford CoreNLP \citep{manning-EtAl:2014:P14-5}:
tokenization, sentence splitting, part-of-speech tagging, lemmatization, named entity recognition, constituency parsing, dependency parsing, coreference resolution, and sentiment analysis.
Due to the format of the user utterances (lowercase with no punctuation), we use caseless models\footnote{\url{https://stanfordnlp.github.io/CoreNLP/caseless.html}} for part-of-speech tagging, constituency parsing and named entity recognition.
We use these annotations for certain handwritten NLU operations.

\subsection{Dialogue Act Classifier}
\label{sec:dialogue_act_classifier}

\begin{table}[th]
\small
\begin{center}
 \begin{tabularx}{\linewidth}{Xccc} 
 \hline
 \textbf{Training Regime} & \textbf{Silver} & \textbf{Gold}  & \textbf{Test F1} \\ 
 \hline
Baseline & 0 & 0 & 0.53 \\
Self-training ($\tau = .95$) & 41,152 & 0 & 0.54 \\
Self-training ($\tau = .75$) & 62,150 & 0 & 0.54 \\
Hand-labeled & 0 & 2,407 & \textbf{0.81} \\
 \hline
\end{tabularx}
\end{center}
\caption{Performance of our Dialogue Act model under different training regimes. All models have access to $10,090$ examples in the MIDAS training set, but training a baseline model solely on these examples suffers from domain shift. \textit{Self-training}, which first uses this baseline model to silver-label a large number of unlabeled \Chirpy examples with confidence above some cutoff $\tau$, then retrains on the union of the two, does not improve performance. \textit{Hand-labelling} a small amount of additional data significantly improves performance. }
\label{da-performance}
\end{table}

Dialogue acts, an ontology over user intents \citep{stolcke2000dialogue, jurafsky1997switch}, have been successfully employed in open-domain dialogue agents \citep{yu2019gunrock}.
We modify MIDAS \citep{yu2019midas}---an annotation schema designed specifically for human-chatbot dialogue---
to better fit the needs of our bot, removing 4 labels\footnote{\textit{apology}, \textit{apology-response}, \textit{other}, and \textit{thanks}} due to low frequency in our conversations and creating 5 new labels: \textit{correction}, \textit{clarification}, \textit{uncertain}, \textit{non-compliant}, and \textit{personal question}.
In total, our modified schema has 24 labels.

Evaluated on the MIDAS test set, a fine-tuned BERT baseline achieves .78 micro-F1; however, evaluated on an OOD test set composed of our own conversations, it achieves only .53 (Table \ref{da-performance}).
Although self-training \citep{mcclosky-etal-2006-effective} proved ineffective,
hand-labeling additional OOD conversations achieved a micro-F1 of 0.81.
The predictions of this final model inform navigation, as well as RG-specific NLU.

\subsection{Question Classifier}
\label{sec:app-annotator-question}
Users often spontaneously ask factual questions, personal questions, follow-up questions, and even questions unrelated to the current topic. 
Recognizing and answering these questions is important, particularly for user initiative, but is also non-trivial, as ASR-transcribed user utterances do not contain punctuation.
To recognize questions, we fine-tuned a RoBERTa model \citep{roberta, huggingface} on an simplified version of the Dialogue Act training data, framing the task as binary classification, conditioned only on the user utterance.
This model achieved an F1-score of 0.92 and improved the reliability of question detection. 

\subsection{QA Annotator}
\label{sec:app-annotator-qa}
The \textbf{QA annotator}, an ELECTRA-Large model \citep{clark2020electra} pretrained on SQuAD2.0 \citep{rajpurkar2018know}, performs question answering for the \textsc{News} (\secref{sec:rg-news}) and \textsc{Wiki} (\secref{sec:rg-wiki}) RGs. Unlike other annotators, this annotator does not run unless called by these RGs.

\section{Neural Generation}
\label{sec:app-neural-gen}
Our neural agent is a distilled \cite{hinton2015distilling} version of BlenderBot-3B~\citep{roller2020recipes}, an autoregressive Seq2Seq model trained on Blended Skill Talk~\citep{smith2020together}, Wizard of Wikipedia~\citep{dinan2019wizard}, \mbox{ConvAI2}~\citep{dinan2019second}, and Empathetic Dialogues~\citep{rashkin2019towards}. We distill using \citet{sanh2019distilbert}'s method (as implemented in ParlAI; \citealp{miller-etal-2017-parlai}), using  Adafactor \citep{shazeer2018adafactor} with learning rate $6.25 \times 10^{-5}$, validation loss-based LR reduction, warmup, and FP16 \cite{gupta2015deep}. 
We used a batch size of 1 for training on a single V100 GPU.

For decoding, we use top-k sampling ($k=5$) with temperature $T=0.7$. To encourage response diversity across the conversation, we sample sequences of minimum length randomly chosen from 5, 10, 15, 20, 25; in practice, the length of the generations is 0-2 tokens above the minimum selected length. 
Additionally, we use delayed beam search \citep{massarelli-etal-2020-decoding}, with the conversational history up to 128 tokens in the past serving as context.
After decoding, we first filter out offensive, null, and repetitive responses, as well as questions after the first turn.
We then select a final response based on the posterior likelihood, among other metrics.

\subsection{Analysis}
\label{sec:app-neural-lim}
We find that our model qualitatively outperforms a GPT-2 \cite{radford2019language} baseline fine-tuned on Empathetic Dialogues (\tabref{tab:neural_examples}), with similar latency.
That said, our model still suffers certain limitations out-of-the-box; we discuss strategies for mitigating these issues. 

\paragraph{Diversity-coherence tradeoff} 
For our model, beam search decoding yields coherent but non-diverse responses, while stochastic decoding results in nonsensical generations even under top-\textit{p}~\citep{holtzman2019curious} or top-\textit{k}~\citep{fan2018hierarchical} sampling. 
Delayed beam search, which samples the first few tokens before defaulting to beam search, yielded more stable behavior than stochastic decoding, and better diversity than beam search.

\paragraph{Degeneration} The model outputs conversation-ending phrases (e.g., ``I have to go'', ``It was nice talking'') after 7 turns, hurting user experience. 
Manual examination of the training data revealed this is due to the short lengths of conversations in the training data collected via crowdworkers. 
We alleviated this with a manual blacklist of conversation-ending phrases and forcing \textsc{Neural Chat} (\secref{sec:rg-neuralchat}) to hand off to another RG prior to the 7-turn limit. 

\paragraph{Changing the topic} Our model tends to generate one or more follow-up questions in every utterance, which are often unrelated to the current topic, creating a disorienting topic change (e.g. \fq{What's your favorite color?}). To address this, \textsc{Neural Chat} limits the number of questions in each neural chat \textit{conversation} to one, with subsequent questions truncated. 
During conditional generation, questions are deleted altogether.

\paragraph{Latency} Generating responses can take up to 2s in the worst-case, averaging just under 1s. 
To mitigate this, we \textbf{pre-fetch} responses from DistillBlender at the start of every turn, instead of waiting for the Neural Chat RG to request a neural response. If the responding RG returns a response without utilizing the DistillBlender response during the turn, we ignore its output and return early. 
Although this results in unused pings to the neural model, we find that the trade-off is worth it.

\begin{table*}[ht]
\small
    \centering
    \begin{tabular}{p{0.2\textwidth}p{0.2\textwidth}p{0.23\textwidth}p{0.23\textwidth}}
    \toprule
        \centering \textbf{Prev. bot utterance} & \centering \textbf{User utterance} &  \centering\textbf{Blender (distilled)} & \centering \textbf{GPT-2 baseline} \tabularnewline
    \midrule
        I love pizza, what's your favorite toppings? & Pepperoni. & That's a good choice. Do you have a favorite kind of cheese? Mine is mozzarella. & Oh yea I love that pizza too. I don't like too much other toppings. \\
    \midrule 
        What did you do today? & Alexa, what's up? & Nothing much. I just got back from a run. I'm exhausted. & I'm sorry to hear that. What happened? \\
    \midrule
        Have you been doing anything productive? & Well, I've been trying to be productive but I keep on breaking things. & Breaking things can be very frustrating. I hope you are able to get it all sorted out. & That's no good, it's not productive to be stuck on something and not get to something else. \\
    \bottomrule
    \end{tabular}
    \vspace{5mm}
\caption{\small Responses from our neural generator, distilled from BlenderBot-3B (punctuation and capitalization added for clarity), compared to a GPT-2 baseline fine-tuned on Empathetic Dialogues. We find that our neural generator provides stronger performance with similar latency.}
\label{tab:neural_examples}
\end{table*}

\section{Additional RG Details}
\label{sec:app-addl-rgs}
\subsection{Launch}
\label{sec:app-rg-launch}
At the beginning of the conversation, the bot initially selected icebreakers at random. However, we eventually found that certain icebreakers tended to fare better than others. Specifically, conversations starting with food-related icebreakers (e.g. ``Do you have any recommendations for what I should cook at home?'') had an average rating was 3.49 over a sample of 1405 conversations, compared to an average rating of 3.43 for non-food-related icebreakers (e.g. ``What did you do over the weekend?'') over a sample of 1418 conversations. Digging deeper, we found that if the second turn is handled by the Food RG, we achieved an average rating of 3.64 over 606 conversations, compared to an average rating of 3.49 if the second turn is handled by the Neural Chat RG, over 1684 conversations (second turns are mainly handled by Food and Neural Chat RG's, but sometimes by others).

This prompted us to update our Launch RG so that we open with a food-related question for all conversations, hence increasing the frequency of handing over to the Food RG.

\subsection{News}
\label{sec:app-rg-news}
The \bfsc{News} RG (\secref{sec:rg-news}) curates global news from The Washington Post\footnote{\url{https://washingtonpost.com}} and The Guardian\footnote{\url{https://theguardian.com}}.
Article titles, topic categories, body texts, dates, and content URLs are stored in a constantly updating ElasticSearch index. 
When a topic or entity available in our index appears in conversation, the News RG brings up related stories from our database. In addition, \textsc{News} also initiaties conversations about currently trending news topics by scraping trending news from Google Trends\footnote{\url{https://trends.google.com}}.

\paragraph{Behavior} To produce a prompt usable in conversation, we rephrase the headline to conversational form using GPT-3 davinci-instruct-beta.\footnote{We use the following prompt: ``Paraphrase news headlines into a complete, grammatical sentence in plain English. The sentence should be in the past tense.''}
If the user expresses interest in continuing the conversation, the we provides a conversational summary generated by Pegasus-Multinews \citep{zhang2019pegasus, fabbri-etal-2019-multi}. Summaries are decoded using 8 beams and a maximum of $50$ tokens for conversationality, and are pre-generated for efficiency; if the neural module fails, we instead use an extractive summary \citep{mihalcea-tarau-2004-textrank}.

\paragraph{Follow-up} If the user continues to be engaged, we prompt for questions or comments. 
If a comment is detected, a neural response is generated using a set of hand-written prefixes; 
If a question is detected (\ref{sec:app-annotator-question}), they are answered via the QA annotator (\ref{sec:app-annotator-qa}).
We then conversationally paraphrase the answer using a GPT-2-medium model \citep{radford2019language} fine-tuned on Topical Chat \cite{gopalakrishnan2019topical} to produce a more human-like response. 
We use the truncated conversational history as the input history and a merged representation of the answer and the span as the the factual content. 
It outputs a conversational-sounding paraphrase of the answer.
Finally, we rank the generated paraphrases using Fused-PCMI \cite{paranjape2021pcmi}.

\subsection{Wiki}
\label{sec:app-rg-wiki}
To support our goal of high-coverage world knowledge (\secref{sec:intro}), the Wiki RG uses Wikipedia articles as grounding to discuss any entity that interests the user and that is not handled by any other RG.
Our goal is to allow the user to conversationally discover interesting information about the entity.

\subsubsection{Data} 
\label{sec:app-wiki-data}
We use the Wikipedia dump from May 20th, 2020\footnote{\url{https://dumps.wikimedia.org/backup-index.html}}, processed using MWParserFromHell\footnote{\url{https://mwparserfromhell.readthedocs.io/en/latest}} and Spark.\footnote{\url{https://spark.apache.org}}
We store our data in a large ElasticSearch index.

\begin{figure*}[ht]
\centering
\vspace*{-3mm}
\includegraphics[width=\textwidth]{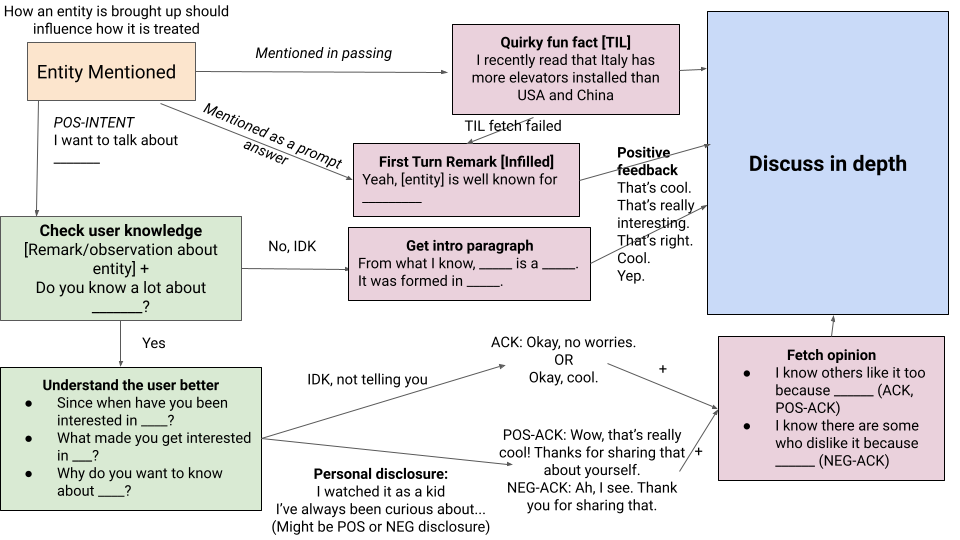}
\caption{The Wiki RG conversational flow: possible user responses are captured in the edge labels, while bot responses are represented by the vertices.}
\vspace*{-3mm}
\label{fig:wiki_flow}
\end{figure*}

\subsubsection{Behavior} 
\label{sec:app-wiki-behavior}
Wiki RG facilitates a discussion about an entity based on how it came up in conversation (see Fig.~\ref{fig:wiki_flow}). 
If the user initiates an discussion about an entity, the RG encourages the user to share their own knowledge and experience about the entity. 
Otherwise, if the entity came up only in passing or as a response to a bot prompt (e.g. ``What's a country you would like to visit?"), then the RG responds with an `infilled' remark (discussed below) or an interesting fact (i.e. `TILs' scraped from the /r/todayilearned subreddit) about the entity. 
These conversation starters serve the purpose of drawing the user into a more conversational dialog about the entity before proceeding to a more content-rich discussion of it.

\begin{figure*}[ht]
\centering
\vspace*{-3mm}
\includegraphics[width=0.8\textwidth]{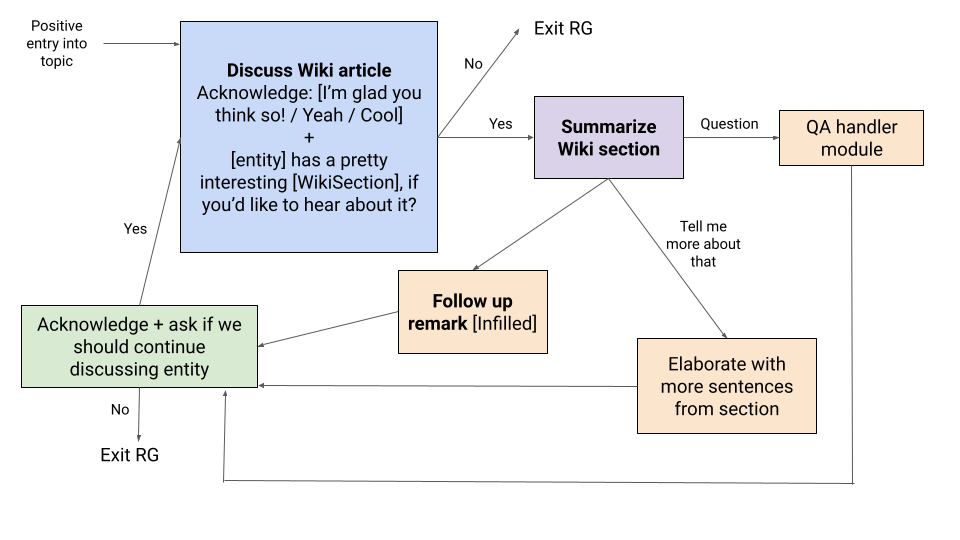}
\caption{The Wiki RG ``Discuss in depth" conversational loop}
\vspace*{-3mm}
\label{fig:wiki_discuss_loop}
\end{figure*}

\paragraph{Discussing the entity in depth.} If the user responds positively to our initial discussion of the entity, we begin a ``Discuss in depth" conversation loop (see Fig.~\ref{fig:wiki_discuss_loop}). 
Our bot provides a summary of some section of the entity's Wikipedia article and handles the user's sentiments, opinions, and questions appropriately before checking if the user would like to continue with the discussion. 
If the user responds affirmatively, we suggest another section for discussion, otherwise we exit the RG. 
This setup ensures that the user is not overly fatigued by the amount of information generated in these section summaries, while allowing interested users to discuss engrossing topics in great depth.

A short example Wiki interaction is shown in Turns 6 through 10 of Table \ref{table:example_dialogue}. 

\subsubsection{Template-Based Infilling}
\label{sec:app-wiki-infilling}
To provide the user with rich, coherent conversation for a wide class of entities, we developed a novel method---\textit{infilling}---which generates interesting remarks from handwritten templates based on relevant context.  For example, given the actor Keanu Reaves as the current entity, the template \textit{I love how \tm{actor} acted in \tm{film}, especially their \mask} might be infilled as follows: \textit{I love how \tm{Keanu Reeves} acted in \tm{The Matrix}, especially their \maskr{ability to freeze time}}. By defining a diverse set of templates for each entity category, we are able to provide expressive yet controllable conversation on many different types of entities. In effect, this acts as a more flexible version of standard slot-filling methods that does not require a structured knowledge base.

Infilling has the following steps: 
\begin{itemize}
    \item A set of templates and appropriate contexts is \textbf{retrieved}. Given some entity, we select a set of handwritten templates based on its Wikidata category (e.g. \textit{actor}, \textit{musical instrument}).
    For each template, we retrieve an appropriate short context from Wikipedia (approximately 3 sentences) using the mean-pooled GloVe-based method of \cite{arora2016simple}.
    \item Given each (context, template) pair, an \textbf{infiller} model fills in the blanks. This is parameterized by a BART-base model trained on a dataset generated by $\sim 4300$ examples, mostly generated using GPT-3 \citep{brown2020language} and augmented by hand-written examples.
    \item The infills are \textbf{reranked} by an aggregate DialogRPT \citep{gao2020dialogue} and likelihood score as measured by a GPT-2-medium model fine-tuned on Empathetic Dialogues.
\end{itemize}

\subsubsection{TIL's: Conversational Paraphrasing}
\label{sec:app-wiki-tils}
We use this RG as a testbed for our conversational paraphrasing system. 
The system takes as input the truncated conversational history, and some knowledge context (either a TIL about the current entity, or an excerpt of the Wikipedia article, selected based on TF-IDF similarity to the user's response to an open-ended question). 
It outputs a conversational-sounding paraphrase of the knowledge context. 
The model was trained by finetuning a GPT-2-medium language model \citep{radford2019language} on a processed and filtered version of the TopicalChat dataset \citep{gopalakrishnan2019topical}. 
The paraphrases are generated using top-$p$ decoding with $p=0.75$ and temperature $\tau=0.9$, and we pick the one which has the highest unigram overlap with the knowledge context.

\subsection{Opinion}
\label{sec:app-rg-opinion}

Exchanging opinions is a core part of social chit-chat. 
To form a stronger sense of personality, and to seem more relatable, it is important that our bot can also express its opinions.
The Opinion RG's goal is to listen to users' opinions on certain topics, and reciprocate with its `own' opinions (sourced from Twitter) on those topics. 

\subsubsection{Data} 
To collect both positive and negative opinions, we queried a Twitter stream\footnote{https://developer.twitter.com/en/docs/tutorials/consuming-streaming-data} using a regex to collect tweets of the form \texttt{``i (love|like|admire|adore|hate|don't like|dislike) TOPIC because REASON''}, where \texttt{TOPIC} and \texttt{REASON} can be any text.
We collected ~900,000 tweets, which are stored on a Postgres table hosted on AWS Relational Database Service (RDS).
Of these, we manually whitelisted 1012 reasons across 109 popular topics. 
To avoid speaking inappropriately about sensitive topics, we only whitelist uncontroversial entities (such as animals, foods, books/movies/games, everyday experiences such as working from home, being sick, days of the week, etc.), and ensured that all reasons, including negative ones, are inoffensive and good-spirited.

\subsubsection{Behavior}
Currently, the Opinion RG activates when the user mentions one of the whitelisted entities (e.g. \tabref{table:example_dialogue}, Turn 8).
We ask whether the user likes the entity and classify their response using the CoreNLP sentiment classifier (\secref{sec:corenlp}). 
We then either agree or disagree with the user. 
If we disagree, we either ask the user for their reason for their opinion, or supply a reason why we disagree, and ask what they think of our reason.
Ultimately, we want the user to have a positive experience with our bot, so regardless of whether we disagree or agree with the user, we will ask the user their opinion on a related entity, and always agree with the user about the new entity. 
The conversation may end earlier, as we detect on each turn whether the user is still interested via their utterance length.
If the utterance contains less than 4 words, and it does not contain any of the `agreement' words (such as `same', `me too', etc.) we will hand off the conversation to another RG.
Even when the RG is not active, it keeps track of whether the user has already expressed an opinion on an entity, by applying a regex similar to that applied to the tweets.

\subsubsection{Agreement Policies}
\label{sec:app-rg-opinion-agreement}
Disagreement is an unavoidable part of human-human conversations, and we hypothesize that occasional disagreement is necessary in order for our bot to have a convincing and individual personality.
To test this, we implemented three policies:
\begin{itemize}
    \item 
(i)~\texttt{ALWAYS\_AGREE} -- we always agree with the user's sentiment on the entity;
    \item 
(ii)~\texttt{LISTEN\_FIRST\_DISAGREE} -- first we ask the user's reason for liking/disliking the entity, then we offer our reason for disagreeing with their sentiment; and
    \item 
(iii)~\texttt{CONVINCED\_AGREE} -- we initially disagree with the user's sentiment on the entity, but after the user gives their reason for liking/disliking the entity, we switch our sentiment to match the user's (i.e. we are convinced by the user).
\end{itemize}
To evaluate the policies, we ask the user \textit{Would you like to continue sharing opinions?} and interpret the desire to continue is an indication of a successful policy. 
\tabref{tab:cont_rate_opinion} shows that users prefer \texttt{ALWAYS\_AGREE} and \texttt{LISTEN\_FIRST\_DISAGREE} over \texttt{CONVINCED\_AGREE}, and all policies have high continuation rates, suggesting that disagreement can be a positive and stimulating part of a conversation, but that the manner and delivery of the disagreement is an important factor.

\begin{table}
    \small
    \centering
    \begin{tabular}{lr}
    \hline
     \textbf{Policy Name}                    &   \textbf{Continuation Rate (95\% CI)}\\
    \hline
    \texttt{CONVINCED\_AGREE}            & .527 $\pm$ .0349  \\
     \texttt{ALWAYS\_AGREE}              & .587 $\pm$ .0086 \\
     \texttt{LISTEN\_FIRST\_DISAGREE}    & .587 $\pm$ .0128  \\
    \hline
    \end{tabular}
    \caption{Continuation rate for each agreement policy. 
    The Confidence Intervals (CI) differ due to different sample sizes (ALWAYS\_AGREE receives 0.5 of traffic, LISTEN\_FIRST\_DISAGREE receives 0.3, CONVINCED\_AGREE receives 0.2).}
    \label{tab:cont_rate_opinion}
\end{table}

\subsection{Food}
\label{sec:app-rg-food}
The Food RG also focuses on scripted responses to discuss foods and give suggestions. It is often activated at the beginning of the conversation when Neural Chat RG prompts a user for what they have eaten today. The Food RG then goes through a sequence where it asks the user about their favorite variant of that food (e.g. favorite pizza topping), mentions the bot's favorite variant, and possibly provides a fun fact about the food. The Food RG is backed by food data scraped from Wikipedia structured in such a way that subclasses and variants of food are linked to each other. It also uses templated responses with neural infilling to generate descriptions of foods or comments on what the user likes, allowing for variation and flexibility for more interesting responses.

\subsection{Movies}
\label{sec:app-rg-movies}

The Movies RG is designed to deliver a high-quality scripted conversation about a movie the user specifies, using information drawn from the Alexa Knowledge Graph.\footnote{The Alexa Knowledge Graph is an Amazon-internal resource; our team was given access to parts of it.} Currently, the RG is activated when the user asks to talk about movies, mentions a movie keyword (such as \textit{movies} or \textit{film}) or talks about any movie-related entity (e.g. \textit{Saving Private Ryan}, \textit{Meryl Streep}, \textit{the Coen brothers}, etc.). 
Once activated, the RG typically asks the user to name a movie, asks the user's opinion on it, gives a fun fact about the movie, asks the user their opinion on an actor in the movie, then asks the user if they've seen a different movie featuring that actor (See Turns 4-7 in Table \ref{table:example_dialogue}). The RG uses treelets (\secref{sec:design}) to organize the dialogue graph, hand-written templates to form the bot utterances, and a mixture of regexes and the CoreNLP sentiment classifier (\secref{sec:corenlp}) to classify the user's responses.

\subsection{Music}
\label{sec:app-rg-music}

Similar to the Movies RG, the Music RG is designed to deliver scripted conversations about musical entities that the user specify. The RG is activated when a musician/band or a music keyword (such as \textit{music} or \textit{songs}) is mentioned. Once activated, the Music RG engages in a conversation specific to the type of the musical entity that was mentioned. Unlike the Movies RG, the Music RG has a randomized internal prompting system that allows the conversation to be centered around music even when a scripted conversation is exhausted for a specific entity. For example, after the Music RG goes until the end of a scripted conversation for a musician, it can ask for an internal prompt, and start a conversation about musical instruments, songs, or music in general. The randomized nature of the internal prompting system makes the conversation more flexible, and mitigates some of the weaknesses of scripted conversations mentioned in \secref{sec:app-rg-movies}.

\subsection{Sports}
\label{sec:app-rg-sports}
The Sports RG is designed to deliver up-to-date and high-quality conversations on a sport for which the user expresses interest. 
Currently, we support conversations on NFL football and NBA basketball, the two most-watched sports in the US. 
When prompted to discuss sports, the user is asked if they are a fan of these two sports.
If so, they are asked for their favorite team, but otherwise the conversation moves to a different RG. The RG supports detailed, factual conversation on the user's favorite team, as well as their favorite player on that team. The Sports RG is backed by an ESPN API scraper that pulls information on all NFL and NBA teams (their game schedule, their roster, wins/losses, game analysis, etc.) and facts about all  players (their age, position, college, statistics, and expert analysis on their overall play). For example, if the user is a fan of the Denver Broncos, the RG is capable of discussing the Broncos' most recent game (who won/lost, what the score was, what player played well, etc.) and then transitions into discussing a specific Broncos player from the game that the user likes. By utilizing automatic summarization, we are able to intersperse current, specific analysis of their favorite player or team that comes directly from ESPN analysts, giving the conversation a sophisticated and natural tone.

\subsection{Fallback}
\label{sec:neural_fallback}
\label{sec:simple_rgs}
When all other RG's fail to produce a suitable response, we rely upon two fallback RG's that always execute. 
The Neural Fallback RG selects a fallback responses generated by the DistillBlender model (\secref{sec:distillblender}), with all questions removed.
Given that the model is trained on end-to-end dialogue, we find that this is a good conversational baseline.
If the Neural Fallback RG fails, we resort to the Fallback RG, 
which returns a pre-written generic fallback (e.g. ``Sorry, I don't know how to answer that'').

\section{High-Initiative User Handling: Example Dialog Flows}
\label{sec:abrupt}

Our system uses high-initiative handler functions to detect 18 distinct types of one-turn digressions that interrupt the conversational flow. General examples include complaints and clarifying questions. A comprehensive list of high-initiative situations that our bot can handle and corresponding examples of bot responses are provided in Table~\ref{tab:abrupt_initiative}. These functions can be started in any RG, and allow each RG to pause the flow of the current dialogue tree to handle high-initiative behaviors before either returning to the original conversation or changing the subject.

To avoid repetitiveness, high-initiative handlers are generally non-reentrant, and cannot be triggered consecutively. 

\begin{table*}[tp]
    \small
    \centering
    \begin{tabularx}{\textwidth}{XXX}
        \toprule
            \textbf{High-Initiative} & \textbf{User utterance} & \textbf{Possible response}  \\
            \textbf{User Behavior} & &\\
        \midrule
            Was misheard & That's not what I said. & Sorry for the misunderstanding; \\
            & &can you repeat that?\\
            Complains bot was unclear & I don't know what you are talking about. & Oops, let me say that again.\\
            Complains bot  & Alexa, you already told  & You're right, I forgot. Sorry. Let's move \\
            was repetitive &me that. &  on to something else. \newtopic \\
            Complaints about  & That's none of your  & Sorry, maybe that was too personal. \\
            privacy & business. & I'm happy to talk about something  \\
            && else. \newtopic\\
            Asked about  & Do you like the weather? & I live in the cloud so I'm not  \\
            weather && sure what the weather is like on earth!\\
            Said Alexa  & Alexa, play Despacito! & This is an Alexa Prize Socialbot... \\
            command & & I can't execute typical Alexa  \\
            & & commands...But I'd love to keep talking \\
            & &to   you! As I was saying, \prev\\
            Wants name  & Actually, that's not my name. & Oops, it sounds like I got your \\
            correction && name wrong. I'm so sorry about \\
            && that! I won't make that mistake again.\\
            Requested repetition & Alexa, can you say  & Oops. Let me repeat that. \prev \\
            & that again? & \\
            Requested own name & What's my name? & If I remember correctly, your\\
            && name is \texttt{[NAME]} \\
            Got cut off & But--- & Oops, I think my microphone \\
            && stopped working for a second. \\
            && Can you say that again?\\            
            Asked for our age & How old are you? & Hmm, I don't know. I'm  just a \\
            &&bot, and I don't think we have \\
            && birthdays.\\
            Asked clarifying  & Wait, did you  & Yes, that's what I said. \\
            question & say ``elephants?" & \\
            Questioned bot's  & How can you listen to music  & Well, I'm just a bot, but\\
            humanity & if you're a bot? &  I live vicariously through others. \\
            Asked personal question & What do you like? & Thanks for asking! \texttt{[NEURAL]}\\
            Interrupted with question & Wait, I have a question. & Sure, what's up? \\
            Expresses desire to chat & Let's chat! & Okay. What would you like to \\
            && talk about? \\
            Asked for story & Tell me a story. & Here's a story that someone once \\
            && told me. \texttt{[STORY]}\\
            Gave bot compliment & I like talking to you! & Thanks for saying that. I'm still\\
            &&  learning, but it means  a lot to \\
            && hear that from you.\\
        \bottomrule
    \end{tabularx}
    \caption{Excerpts of possible bot responses to 18 high-initiative user behavior modes. \prev\; refers to the previous bot utterance; \newtopic\; refers to a sampled prompt from a new RG. \texttt{[NAME]} is the user's name as obtained in the opening turns, and \texttt{[NEURAL]} refers to a DistillBlender-based random response. \texttt{[STORY]} is a handwritten anecdote, omitted here for brevity.}
    \label{tab:abrupt_initiative}
\end{table*}

\section{Additional Neural Modeling Experiments}
\newcommand{\dsap}[0]{\textbf{AP2021}\xspace}


In addition to BlenderBot-3B, we experimented with a variety of autoregressive and non-autoregressive models for text generation. All models are evaluated on an internal dataset of 517 conversation excerpts from early 2021 where the Neural Chat RG was active \dsap. We perform qualitative evaluation by passing in each conversational excerpt to the model of interest, and comparing the resultant generation(s) with the original neural generation from GPT2ED. We detail results here.

\paragraph{DialoGPT.} DialoGPT \citep{zhang2019dialogpt} is a GPT-Medium model that has been further fine-tuned on a set of Reddit threads serving as conversational corpora. We evaluated this model offline on a set of excerpts in two settings: 1) \textbf{zero-shot} and 2) with fine-tuning on Empathetic Dialogues \textbf{(DialoGPT2ED)}. In the \textbf{zero-shot} setting, the bot responds 18\% of the time with dirty jokes or memetic content unsafe for open-domain conversation on \dsap. After fine-tuning, \textbf{(DialoGPT2ED)} responds almost identically to GPT2ED on \dsap: qualitatively, the lift from DialoGPT2ED is essentially zero. Hence, this system was not deployed.

\paragraph{DistillBART.} DistillBART is our in-house distilled version of BART \cite{lewis2020bart}, a model consisting of a non-autoregressive encoder and an autoregressive decoder, each with 12 layers. Notably, this model has decoding complexity $\mathcal{O}(EN + DN^2)$, where $N$ is the sequence length, and $E, D$ are the sizes of the encoder and decoder stacks, respectively. Following results by~\cite{kasai2020deep} in the domain of neural machine translation, we hypothesized that we could decrease latency while improving performance by decreasing $D$; i.e. removing decoder layers and training the decoder via distillation. We performed DistillBERT-style distillation, distilling a BART-Large fine-tuned on Empathetic Dialogues (BARTED) into versions with 6 \textbf{(DistillBART-6)} and 3 \textbf{(DistillBART-3)} decoder layers. Weight initialization followed a previous setup for BART distillation~\citep{shleifer2020pretrained}. As baselines, we also trained equivalently-sized models without distillation. 

In practice, BART suffered from 1) high latency and 2) mediocre response quality. BART was unable to generate coherent responses stochastically, necessitating the usage of beam search, which hurt decoding speed. On \dsap, average decoding speeds for the 12, 6, and 3 layer models were 894ms, 998ms, and 895ms, showing no significant latency gains, which is attributable to the quadratic dependence within the decoding computation on sequence length; i.e. $N^2 \gg D, E$. Furthermore, while distillation certainly resulted in qualitatively better generations on \dsap than those of non-distilled models, as shown in Table, there was a sharp dropoff in generation quality on all models except the full-sized BARTED teacher. As BARTED was the only usable model, and yielded generations qualitatively similar to GPT2ED, we did not deploy this system.


\end{document}